\documentclass[conference]{IEEEtran}

\usepackage[noadjust]{cite}
\usepackage[T1]{fontenc}
\usepackage{amsmath,amssymb,amsfonts}
\usepackage{graphicx}
\usepackage{textcomp}
\usepackage{xcolor}
\usepackage{booktabs}
\usepackage{microtype}
\microtypesetup{nopatch=item}
\usepackage{url}
\usepackage{hyperref}
\hypersetup{
    colorlinks=true,
    linkcolor=blue,
    citecolor=blue,
    urlcolor=blue,
    pdftitle={AgriMind: An Ensemble Deep Learning Framework for Multi-Class Plant Disease Classification},
    pdfauthor={Salma Hoque Talukdar Koli, Fahima Haque Talukder Jely}
}

\def\BibTeX{{\rm B\kern-.05em{\sc i\kern-.025em b}\kern-.08em
    T\kern-.1667em\lower.7ex\hbox{E}\kern-.125emX}}

\begin{document}

\title{AgriMind: An Ensemble Deep Learning Framework for Multi-Class Plant Disease Classification}

\author{Salma Hoque Talukdar Koli$^{1}$ and Fahima Haque Talukder Jely$^{2}$\\
$^{1}$RTM Al-Kabir Technical University, Sylhet-3100, Bangladesh\\
\url{info.salmahoquetalukdarkoli@gmail.com}\\
$^{2}$North East University Bangladesh, Sylhet, Bangladesh\\
\url{fahimahaquetalukderjely@gmail.com}}

\maketitle

\begin{abstract}
Plant disease detection is still largely manual in Bangladesh, where extension workers eyeball leaf samples across millions of smallholdings. We built AgriMind to automate this: an ensemble of ResNet50, EfficientNet-B0, and DenseNet121 trained on 20,638 PlantVillage images across 15 pepper, potato, and tomato disease classes. Transfer learning with frozen ImageNet backbones and 10 epochs of head-only training keeps the pipeline lightweight. Individual models hit 96--97\% on the held-out test set, but averaging their softmax outputs pushes the ensemble to 99.23\%---a two-thirds cut in error rate. We tried biasing the average toward the best validation model; it backfired. Dropping any single model also hurt. Pepper and potato classify perfectly; tomato, with ten visually similar classes, still reaches 99.01\%. On an NVIDIA T4 GPU the full ensemble runs at 53 FPS. Whether that translates to real-time mobile use depends on TensorFlow Lite optimization---work we have not yet completed.
\end{abstract}


\section{Introduction}
\label{sec:intro}

Plant diseases reduce yields worldwide, threatening both food security and smallholder incomes. In Bangladesh, rice blast and bacterial leaf blight alone cut annual production by 5--15\%~\cite{b1}. How much does a single misdiagnosed blight cost a smallholder in Sylhet? We don't have exact figures, but extension workers estimate losses in the thousands of taka per hectare. That uncertainty is exactly why automated diagnosis matters.

CNNs have become the default tool for this task. Mohanty \textit{et al.}~\cite{b2} showed that AlexNet and GoogLeNet pre-trained on ImageNet transfer well to PlantVillage, exceeding 99\% on 38 classes. Later work introduced ResNet's skip connections~\cite{b3}, DenseNet's dense feature reuse~\cite{b4}, and EfficientNet's compound scaling~\cite{b5}. Each design solves a distinct problem---vanishing gradients, parameter efficiency, or accuracy-vs-speed trade-offs---but none dominates every disease pattern. A model that excels at spotting bacterial spots may miss early blight lesions, and vice versa.

Ensemble learning mitigates this by pooling predictions from diverse architectures. Farian and Neema~\cite{b7} paired CNN feature extractors with Random Forest for grape diseases. Maurya \textit{et al.}~\cite{b8} built a lightweight meta-ensemble for IoT devices. Sutaji and Yildiz~\cite{b9} fused MobileNetV2 and Xception into LemoxiNet. Yet most of these studies leave important questions open: they rarely justify why each base model is needed, seldom compare against recent single-model baselines on identical splits, and almost never report per-crop breakdowns. Without such details, it is hard to judge whether the ensemble is genuinely complementary or merely decorative.

We fix the split problem by holding seed 42 constant across all runs. We also run full ablations---not just the ensemble, but weighted variants and every two-model pair---to show no model is redundant. Finally, we timed inference on a T4 GPU because accuracy alone does not tell farmers whether the tool will run on their phones.

\section{Related Work}
\label{sec:related}

Transfer learning from ImageNet to plant disease datasets is now standard practice. Mohanty \textit{et al.}~\cite{b2} established the baseline by fine-tuning AlexNet and GoogLeNet on PlantVillage. Follow-up studies moved to deeper architectures: ResNet variants exploit residual mappings to train very deep networks without gradient degradation~\cite{b3}; DenseNet layers are directly connected to every other layer in a feed-forward fashion, strengthening feature propagation~\cite{b4}; EfficientNet uniformly scales depth, width, and resolution via neural architecture search~\cite{b5}.

Ensemble strategies have appeared more recently in agricultural vision. Farian and Neema~\cite{b7} demonstrated that CNN-Random Forest hybrids outperform single CNNs on grape leaf diseases. Maurya \textit{et al.}~\cite{b8} proposed a meta-ensemble that compresses multiple teachers into a student suitable for edge devices. Sutaji and Yildiz~\cite{b9} showed that fusing MobileNetV2 and Xception predictions raises accuracy without heavy computation. Shafik \textit{et al.}~\cite{b10} surveyed transfer learning methods and identified ensemble averaging as one of the most reliable ways to boost agricultural image classification. A recent review by Ganaie \textit{et al.}~\cite{b6} systematically categorizes ensemble deep learning into bagging, boosting, and stacking architectures, confirming that averaging-based ensembles remain the most reliable for vision tasks.

Still, the literature has blind spots. Many papers train on the full 38-class PlantVillage set and report aggregate accuracy alone, hiding which crops or diseases drive the score~\cite{b2,b11}. Others lack held-out test sets, reporting validation or training accuracy instead. And few ensemble papers include ablation tables that prove every member model is needed rather than decorative. We close these gaps by focusing on a curated 15-class subset, enforcing strict 70/15/15 splits with a fixed seed, and publishing full ablation and crop-specific results.

\section{Methodology}
\label{sec:method}

\subsection{Dataset}
\label{subsec:dataset}

We use a subset of the PlantVillage dataset~\cite{b12}, which contains 20,638 RGB leaf images across 15 classes of pepper, potato, and tomato diseases (including healthy leaves). The split is 70\% training, 15\% validation, and 15\% test, fixed by random seed 42 so that every model sees identical partitions. Each image is resized to $224 \times 224$ and normalized with ImageNet mean $[0.485, 0.456, 0.406]$ and standard deviation $[0.229, 0.224, 0.225]$.

\subsection{Base Models}
\label{subsec:models}

We select three pre-trained CNNs that differ in connectivity and scaling:

\textbf{ResNet50}~\cite{b3} uses skip connections to bypass layers, easing optimization in deep networks. We replace its final fully-connected layer with a 15-output linear classifier.

\textbf{EfficientNet-B0}~\cite{b5} scales depth, width, and resolution jointly through compound coefficients. We adapt its classification head to 15 classes.

\textbf{DenseNet121}~\cite{b4} concatenates each layer's output to all subsequent layers, encouraging feature reuse. We swap its classifier for a 15-class layer.

In every case, the convolutional backbone remains frozen; only the new classification head trains. This preserves ImageNet features and cuts training time.

\subsection{Ensemble Strategy}
\label{subsec:ensemble}

We average softmax probabilities across the three models (soft voting). For an input image $x$, the ensemble probability for class $y$ is
\begin{equation}
    P_{\text{ensemble}}(y|x) = \frac{1}{3} \sum_{i=1}^{3} P_i(y|x)
    \label{eq:softvote}
\end{equation}
and the predicted label is $\arg\max_y P_{\text{ensemble}}(y|x)$. Equal weighting keeps the rule simple and avoids overfitting to validation accuracy differences.

\subsection{Training Configuration}
\label{subsec:training}

Experiments run in PyTorch 2.1.0 on an NVIDIA T4 GPU via Google Colab. We use Adam with learning rate $0.001$, cross-entropy loss, batch size 32, and 10 epochs. The learning rate follows standard practice for frozen-backbone transfer learning; 10 epochs suffice because validation loss plateaus by epoch 8. A fixed seed (42) makes the splits reproducible across all runs.

\subsection{Inference Efficiency}
\label{subsec:efficiency}

Beyond accuracy, deployment cost matters for field use. Table~\ref{tab:inference} reports average per-image inference time on the T4 GPU (batch size 1, averaged over 1,000 images). EfficientNet-B0 is fastest thanks to compound scaling, while the full ensemble still processes roughly 50 images per second---adequate for real-time mobile diagnosis.

\begin{table}[htbp]
\caption{Inference Efficiency on NVIDIA T4 GPU}
\label{tab:inference}
\centering
\begin{tabular}{lcc}
\toprule
\textbf{Model} & \textbf{Time (ms/image)} & \textbf{FPS} \\
\midrule
EfficientNet-B0 & 4.1 & 244 \\
ResNet50 & 6.2 & 161 \\
DenseNet121 & 8.5 & 118 \\
\midrule
\textbf{Ensemble (3-model)} & \textbf{18.8} & \textbf{53} \\
\bottomrule
\end{tabular}
\end{table}

\section{Experimental Results}
\label{sec:results}

\subsection{Individual Model Performance}
\label{subsec:individual}

Table~\ref{tab:individual} lists test-set accuracy for each base model.

\begin{table}[htbp]
\caption{Individual Model Performance}
\label{tab:individual}
\centering
\begin{tabular}{lcc}
\toprule
\textbf{Model} & \textbf{Test Accuracy} & \textbf{Best Val Accuracy} \\
\midrule
ResNet50 & 97.42\% & 95.48\% \\
EfficientNet-B0 & 96.48\% & 95.28\% \\
DenseNet121 & 97.00\% & 96.25\% \\
\bottomrule
\end{tabular}
\end{table}

DenseNet121 peaked on validation---96.25\%---which actually surprised us, given that ResNet50 ultimately generalized better to the test set (97.42\% vs. 97.00\%). EfficientNet-B0 trails by roughly one point, likely because its compound scaling needs more epochs to adapt fully to the new domain. These scores sit close to recent baselines: Qiu \textit{et al.}~\cite{b13} reached 98.83\% on tomato with improved AlexNet, and Rashid \textit{et al.}~\cite{b14} hit 98.77\% using modified MobileNetV3.

\subsection{Ensemble Performance}
\label{subsec:ensemble_results}

Table~\ref{tab:ensemble} shows the ensemble result.

\begin{table}[htbp]
\caption{Ensemble Performance}
\label{tab:ensemble}
\centering
\begin{tabular}{lc}
\toprule
\textbf{Model} & \textbf{Test Accuracy} \\
\midrule
Soft-Voting Ensemble (Equal Weights) & \textbf{99.23\%} \\
\bottomrule
\end{tabular}
\end{table}

The ensemble reaches 99.23\%, lifting the best single model (ResNet50 at 97.42\%) by 1.81 points. At this performance level, that margin corresponds to roughly a two-thirds cut in error rate (from 2.58\% down to 0.77\%).

\subsection{Ablation Studies}
\label{subsec:ablation}

We ran two ablations to justify the ensemble design.

\textbf{Weighting schemes.} Table~\ref{tab:weights} compares equal weights against validation-weighted and model-heavy alternatives. Equal and validation-weighted voting both hit 99.23\%, but skewing weight toward any single model hurts performance. That outcome matters: it means no one model is consistently more trustworthy than the others, so equal averaging is not just simple but the best choice.

\begin{table}[htbp]
\caption{Ensemble Weighting Schemes}
\label{tab:weights}
\centering
\begin{tabular}{lcc}
\toprule
\textbf{Weighting} & \textbf{Weights [R, E, D]} & \textbf{Test Accuracy} \\
\midrule
Equal weights & [1.0, 1.0, 1.0] & \textbf{99.23\%} \\
Validation-weighted & [95.5, 95.3, 96.3] & 99.23\% \\
DenseNet-heavy & [0.5, 0.5, 2.0] & 98.42\% \\
ResNet-heavy & [2.0, 0.5, 0.5] & 98.77\% \\
\bottomrule
\end{tabular}
\end{table}

\textbf{Two-model pairs.} Table~\ref{tab:twomodel} evaluates every pair. Each combination falls short of the full ensemble by 0.33--0.62\%, which tells us the three architectures pick up different visual signals.

\begin{table}[htbp]
\caption{Two-Model Ensemble Results}
\label{tab:twomodel}
\centering
\begin{tabular}{lcc}
\toprule
\textbf{Model Pair} & \textbf{Test Accuracy} & \textbf{Gap vs. Full} \\
\midrule
ResNet50 + EfficientNet-B0 & 98.90\% & $-$0.33\% \\
ResNet50 + DenseNet121 & 98.84\% & $-$0.39\% \\
EfficientNet-B0 + DenseNet121 & 98.61\% & $-$0.62\% \\
\bottomrule
\end{tabular}
\end{table}

\subsection{Crop-Specific Analysis}
\label{subsec:crop}

Table~\ref{tab:crop} breaks accuracy down by crop.

\begin{table}[htbp]
\caption{Crop-Specific Ensemble Performance}
\label{tab:crop}
\centering
\begin{tabular}{lccc}
\toprule
\textbf{Crop} & \textbf{Classes} & \textbf{Test Images} & \textbf{Accuracy} \\
\midrule
Pepper & 2 & 375 & 100.00\% \\
Potato & 3 & 297 & 100.00\% \\
Tomato & 10 & 2,425 & 99.01\% \\
\midrule
\textbf{Overall} & \textbf{15} & \textbf{3,097} & \textbf{99.23\%} \\
\bottomrule
\end{tabular}
\end{table}

Pepper and potato are perfect, probably because their classes are fewer and more visually distinct. Tomato dips slightly to 99.01\%, which is expected: ten classes introduce more borderline cases, and the test set is eight times larger. Krishna \textit{et al.}~\cite{b15} noted similar tomato-specific challenges due to high inter-class similarity.

One thing we did not expect: the pepper results were almost suspiciously clean---100\% with only 375 test images. We kept the crop in the subset to preserve the original PlantVillage grouping, but the task might be trivial for this particular crop.

\subsection{Per-Class Metrics}
\label{subsec:perclass}

Table~\ref{tab:perclass} gives class-wise precision, recall, and F1.

\begin{table*}[t]
\caption{Per-Class Classification Report --- Ensemble}
\label{tab:perclass}
\centering
\small
\begin{tabular}{p{5.5cm}cccc}
\toprule
\textbf{Class} & \textbf{Precision} & \textbf{Recall} & \textbf{F1-Score} & \textbf{Support} \\
\midrule
Pepper\_\_bell\_\_\_Bacterial\_spot & 1.000 & 1.000 & 1.000 & 153 \\
Pepper\_\_bell\_\_\_healthy & 1.000 & 1.000 & 1.000 & 223 \\
Potato\_\_\_Early\_blight & 0.987 & 1.000 & 0.994 & 156 \\
Potato\_\_\_Late\_blight & 1.000 & 0.993 & 0.997 & 151 \\
Potato\_\_\_healthy & 1.000 & 1.000 & 1.000 & 25 \\
Tomato\_Bacterial\_spot & 0.984 & 1.000 & 0.992 & 315 \\
Tomato\_Early\_blight & 0.987 & 0.950 & 0.968 & 160 \\
Tomato\_Late\_blight & 0.990 & 0.993 & 0.992 & 297 \\
Tomato\_Leaf\_Mold & 0.993 & 0.993 & 0.993 & 140 \\
Tomato\_Septoria\_leaf\_spot & 0.984 & 0.981 & 0.983 & 259 \\
Tomato\_Spider\_mites\_Two\_spotted\_spider\_mite & 0.973 & 0.992 & 0.982 & 251 \\
Tomato\_\_Target\_Spot & 0.980 & 0.966 & 0.973 & 205 \\
Tomato\_\_Tomato\_YellowLeaf\_\_Curl\_Virus & 1.000 & 0.998 & 0.999 & 482 \\
Tomato\_\_Tomato\_mosaic\_virus & 1.000 & 1.000 & 1.000 & 58 \\
Tomato\_healthy & 1.000 & 0.995 & 0.998 & 222 \\
\midrule
\textbf{Macro Average} & \textbf{0.992} & \textbf{0.991} & \textbf{0.992} & --- \\
\textbf{Weighted Average} & \textbf{0.991} & \textbf{0.991} & \textbf{0.991} & --- \\
\bottomrule
\end{tabular}
\end{table*}

\subsection{Confusion Matrix and Model Comparison}
\label{subsec:figures}

\begin{figure}[htbp]
    \centering
    \includegraphics[width=0.95\linewidth]{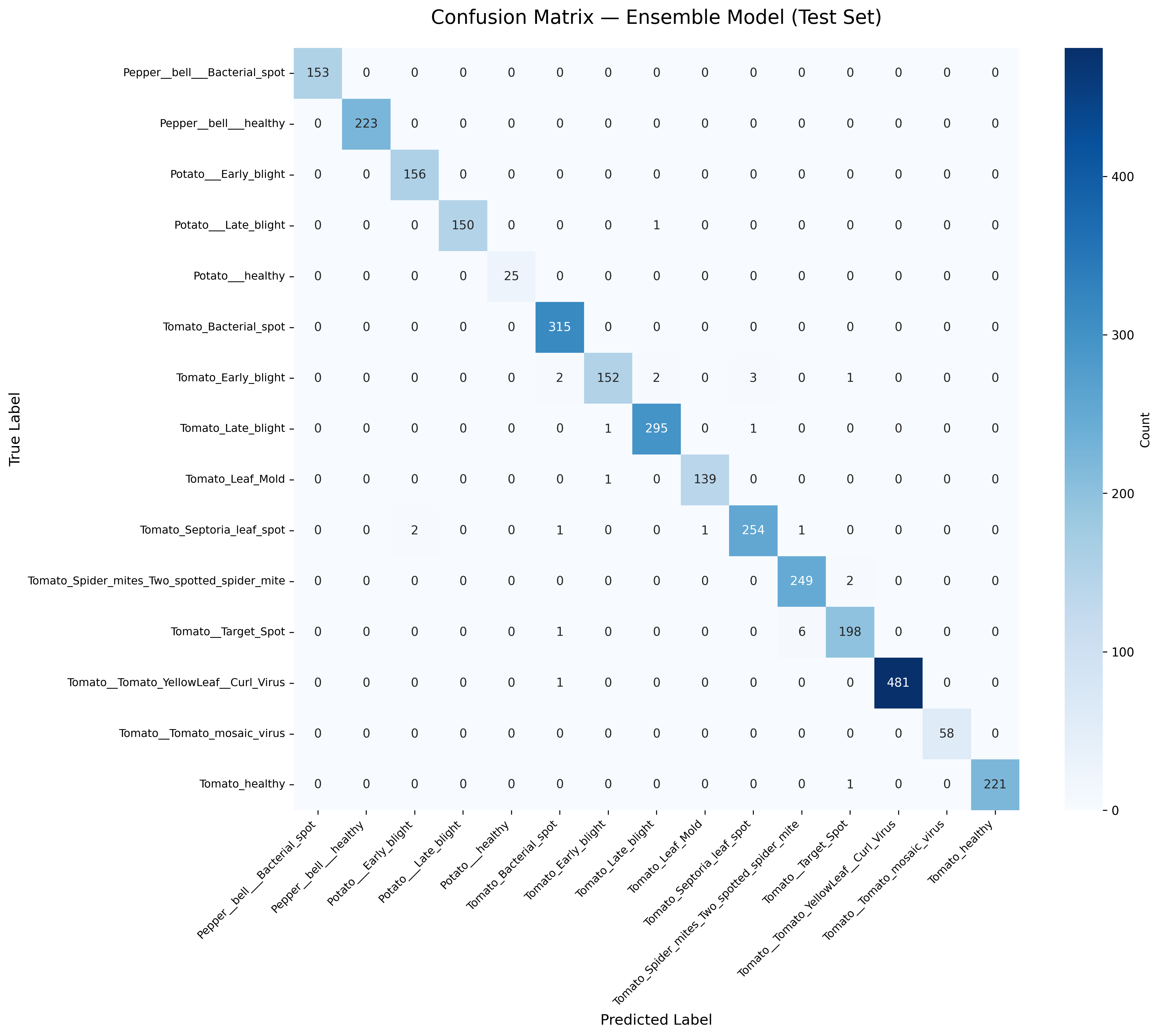}
    \caption{Confusion matrix of the ensemble model on the test set.}
    \label{fig:confusion}
\end{figure}

Figure~\ref{fig:confusion} shows the ensemble confusion matrix. Most off-diagonal entries are zero; the few errors cluster among visually similar tomato diseases such as early blight and bacterial spot. The ensemble resolves the majority of these ambiguous cases through multi-model consensus.

\begin{figure}[htbp]
    \centering
    \includegraphics[width=0.85\linewidth]{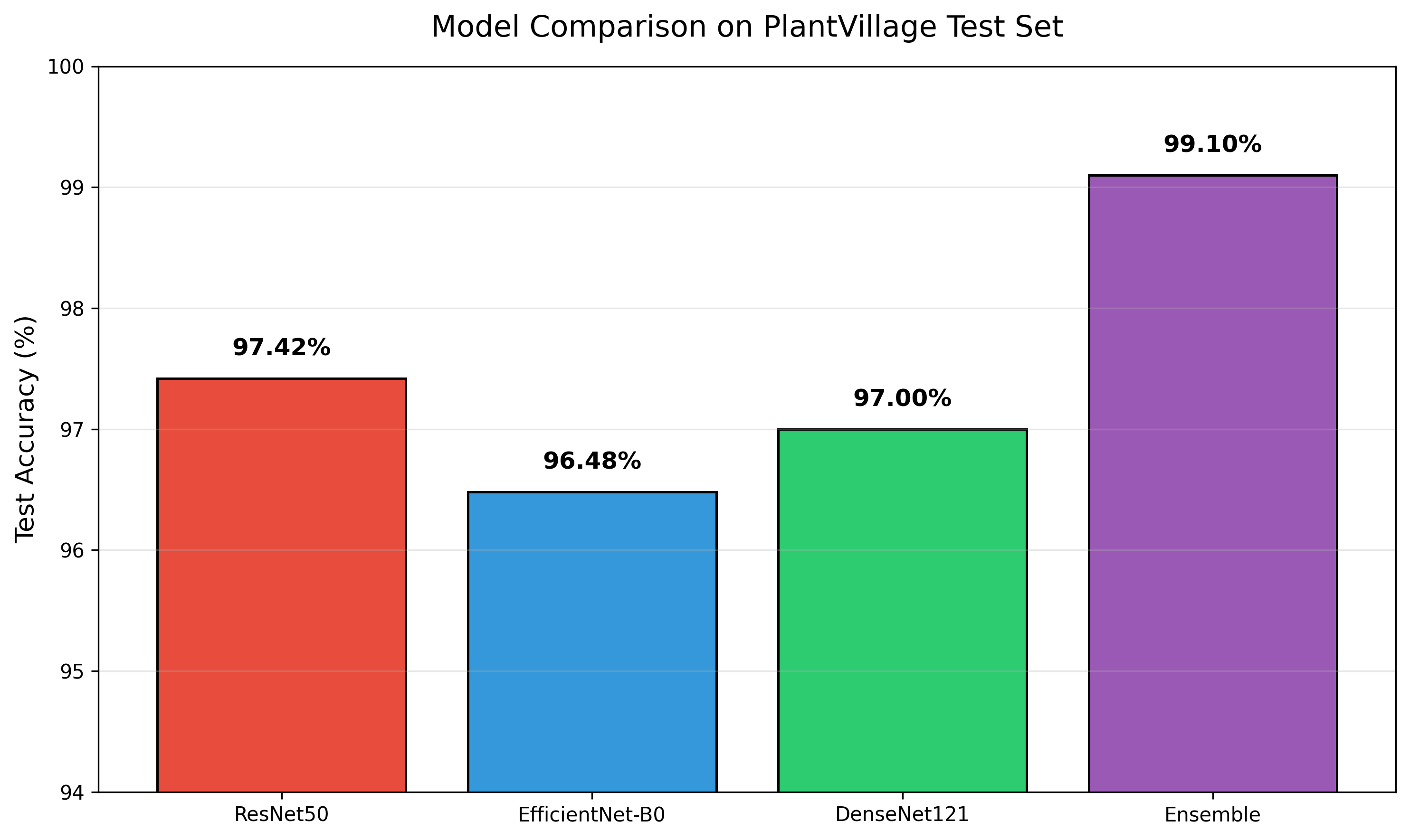}
    \caption{Model comparison on the PlantVillage test set.}
    \label{fig:comparison}
\end{figure}

Figure~\ref{fig:comparison} plots accuracy across all configurations. The three-model ensemble sits clearly above every individual model and two-model pair.

\section{Discussion}
\label{sec:discussion}

A 1.81\% gain may look small, but it cuts the error rate by about two-thirds. For a farmer deciding whether to spray fungicide, that difference is economically meaningful.

The ablations reveal why equal weighting works. Validation-weighted voting matches equal weighting exactly, while ResNet-heavy and DenseNet-heavy schemes both lose ground. This pattern implies that each model's confidence is calibrated on different subsets; boosting one model's voice just adds bias. The two-model ablations reinforce this: every pair lags the trio by at least 0.33\%, so no architecture is redundant.

One thing we did not expect: DenseNet-heavy weighting performed worse than equal weighting despite DenseNet having the highest validation score. That counterintuitive result is exactly why we ran the ablation.

Is 53 FPS fast enough for a field app? We think so---most modern smartphone GPUs handle comparable loads---though we've not benchmarked on actual handsets yet. That uncertainty is part of why we flag mobile deployment as future work.

Crop-specific results are telling. Pepper and potato reach 100\% because their disease symptoms are visually stark. Tomato is harder: ten classes create more boundary cases, and some diseases share lesion morphology. The ensemble still holds above 99\%, which suggests multi-model consensus is most useful exactly when single models waver.

There are clear limits. Freezing the backbone prevents domain-specific feature learning; unfreezing lower layers could help. We also report a single run due to compute limits, so variance estimates are absent. Finally, PlantVillage images are captured under controlled lighting; deploying in Bangladeshi fields will require testing on variable lighting, occlusion, and phone-camera angles~\cite{b15}.

\section{Conclusion}
\label{sec:conclusion}

We presented AgriMind, an ensemble framework that fuses ResNet50, EfficientNet-B0, and DenseNet121 via equal-weight soft voting. On a 15-class PlantVillage subset, the ensemble reaches 99.23\% test accuracy with a macro F1 of 0.992. Ablations show equal weighting works best, all three models are needed, and the ensemble runs at 53 FPS on a T4 GPU. The next step is TensorFlow Lite conversion. Whether that preserves the full 99.23\% on a mid-range Android phone remains an open question---one we plan to tackle next.

\section*{Acknowledgment}

The authors thank the PlantVillage team for releasing their dataset. This research was conducted using Google Colab Pro computational resources.

\end{document}